\documentclass[11pt,letterpaper]{article}
\usepackage{acl2012}
\usepackage{times}
\usepackage{url}
\usepackage{latexsym}
\usepackage{amsmath}
\usepackage{amssymb}
\usepackage{graphicx}
\usepackage{todonotes}
\usepackage{footnote}
\usepackage{multirow}

\makeatletter
\newcommand{\@BIBLABEL}{\@emptybiblabel}
\newcommand{\@emptybiblabel}[1]{}
\makeatother
\usepackage[hidelinks]{hyperref}

\title{Named Entity Recognition with Bidirectional LSTM-CNNs}

\author{
  Jason P.C. Chiu \\
  University of British Columbia \\
  {\tt jsonchiu@gmail.com} \\ \And
  Eric Nichols \\
  Honda Research Institute Japan Co.,Ltd. \\
  {\tt e.nichols@jp.honda-ri.com} \\
}

\date{}

\begin{document}
\maketitle
\begin{abstract}



Named entity recognition is a challenging task that has traditionally required large amounts of knowledge in the form of feature engineering and lexicons to achieve high performance. In this paper, we present a novel neural network architecture that automatically detects word- and character-level features using a hybrid bidirectional LSTM and CNN architecture, eliminating the need for most feature engineering. We also propose a novel method of encoding partial lexicon matches in neural networks and compare it to existing approaches.
Extensive evaluation shows that, given only tokenized text and publicly available word embeddings, our system is competitive on the CoNLL-2003 dataset and surpasses the previously reported state of the art performance on the OntoNotes 5.0 dataset by 2.13 F1 points. By using two lexicons constructed from publicly-available sources, we establish new state of the art performance with an F1 score of 91.62 on CoNLL-2003 and 86.28 on OntoNotes, surpassing systems that employ heavy feature engineering, proprietary lexicons, and rich entity linking information.

\end{abstract}

\section{Introduction}

Named entity recognition is an important task in NLP. High performance approaches have been dominated by applying CRF, SVM, or perceptron models to hand-crafted features \cite{ratinov2009,passos2014,luo2015}. However, Collobert et al. \shortcite{collobert2011} proposed an effective neural network model that requires little feature engineering and instead learns important features from word embeddings trained on large quantities of unlabelled text -- an approach made possible by recent advancements in unsupervised learning of word embeddings on massive amounts of data \cite{collobert2008,mikolov2013} and neural network training algorithms permitting deep architectures \cite{rumelhart1986}. 

Unfortunately there are many limitations to the model proposed by Collobert et al. \shortcite{collobert2011}. First, it uses a simple feed-forward neural network, which restricts the use of context to a fixed sized window around each word \--- an approach that discards useful long-distance relations between words. Second, by depending solely on word embeddings, it is unable to exploit explicit character level features such as prefix and suffix, which could be useful especially with rare words where word embeddings are poorly trained. We seek to address these issues by proposing a more powerful neural network model.

A well-studied solution for a neural network to process variable length input and have long term memory is the recurrent neural network (RNN) \cite{goller1996}. Recently, RNNs have shown great success in diverse NLP tasks such as speech recognition \cite{graves2013}, machine translation \cite{cho2014}, and language modeling \cite{mikolov2011rnnlm}. The long-short term memory (LSTM) unit with the forget gate allows highly non-trivial long-distance dependencies to be easily learned \cite{gers2000}. For sequential labelling tasks such as NER and speech recognition, a bi-directional LSTM model can take into account an effectively infinite amount of context on both sides of a word and eliminates the problem of limited context that applies to any feed-forward model \cite{graves2013}. While LSTMs have been studied in the past for the NER task by Hammerton \shortcite{hammerton2003}, the lack of computational power (which led to the use of very small models) 
and quality word embeddings limited their effectiveness. 

\begin{figure}[t]
\hspace{0.5mm}
\includegraphics[scale=0.72]{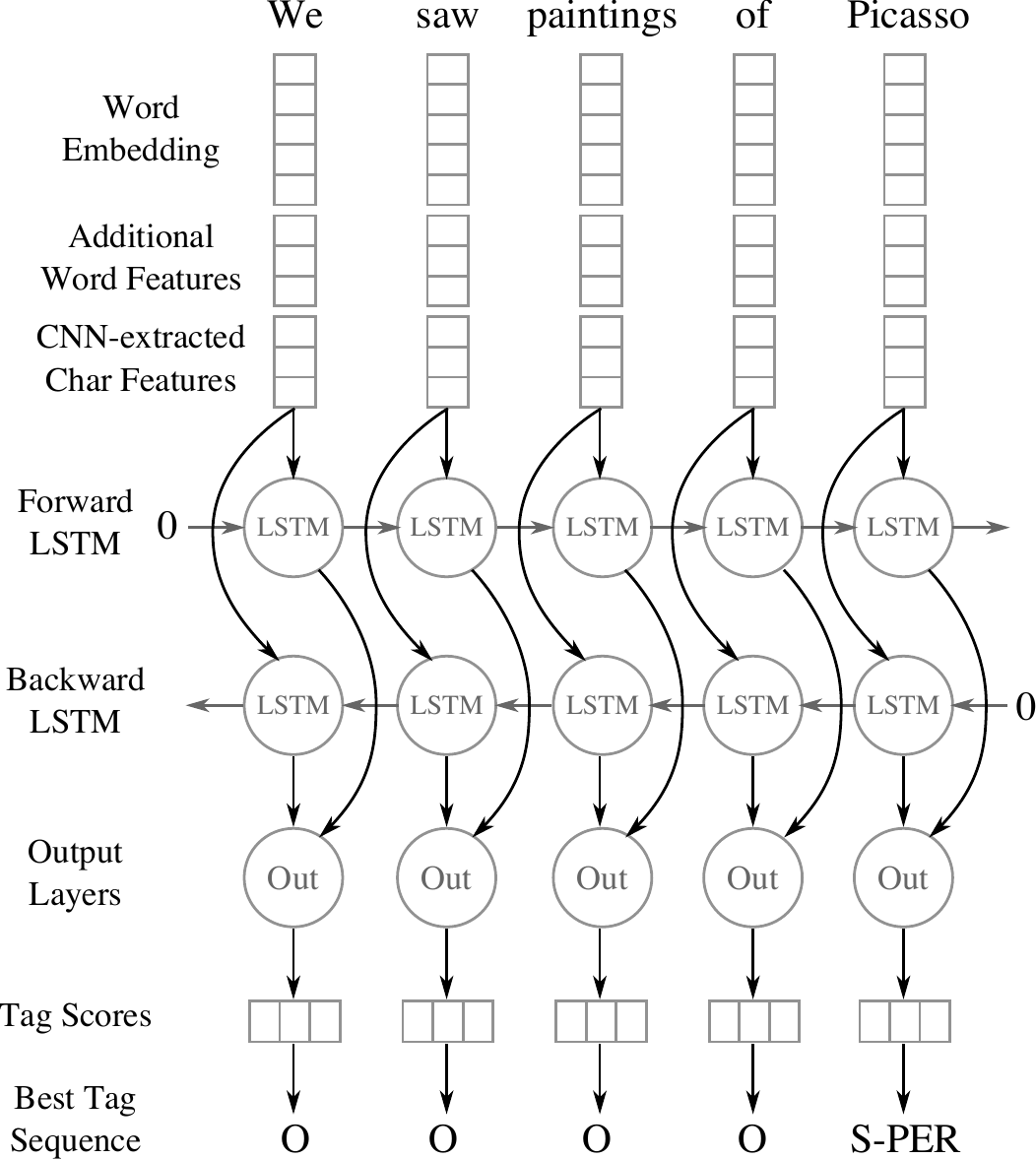} 
\caption{The (unrolled) BLSTM for tagging named entities. Multiple tables look up word-level feature vectors. The CNN (Figure \ref{fig:network-cnn}) extracts a fixed length feature vector from character-level features. For each word, these vectors are concatenated and fed to the BLSTM network and then to the output layers (Figure \ref{fig:network-out}). }
\label{fig:network-main}
\end{figure}

Convolutional neural networks (CNN) have also been investigated for modeling character-level information, among other NLP tasks. Santos et al. \shortcite{dos2015} and Labeau et al. \shortcite{labeau2015} successfully employed CNNs to extract character-level features for use in NER and POS-tagging respectively. Collobert et al. \shortcite{collobert2011} also applied CNNs to semantic role labeling, and variants of the architecture have been applied to parsing and other tasks requiring tree structures \cite{blunsom2014}. However, the effectiveness of character-level CNNs has not been evaluated for English NER. While we considered using character-level bi-directional LSTMs, which was recently proposed by Ling et al. \shortcite{ling2015} for POS-tagging, preliminary evaluation shows that it does not perform significantly better than CNNs while being more computationally expensive to train.

Our main contribution lies in combining these neural network models for the NER task. We present a hybrid model of bi-directional LSTMs and CNNs that learns both character- and word-level features, presenting the first evaluation of such an architecture on well-established English language evaluation datasets. Furthermore, as lexicons are crucial to NER performance, we propose a new lexicon encoding scheme and matching algorithm that can make use of partial matches, and we compare it to the simpler approach of Collobert et al. \shortcite{collobert2011}. Extensive evaluation shows that our proposed method establishes a new state of the art on both the CoNLL-2003 NER shared task and the OntoNotes 5.0 datasets.

\section{Model}

\begin{figure}[t]
\hspace{2mm}
\includegraphics[scale=0.83]{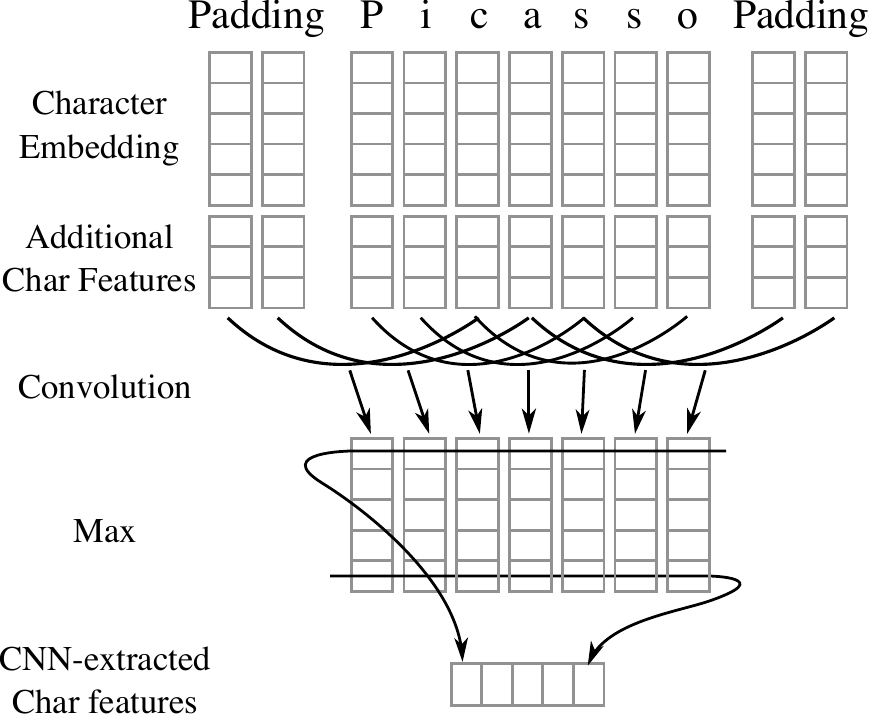} 
\caption{The convolutional neural network extracts character features from each word. The character embedding and (optionally) the character type feature vector are computed through lookup tables. Then, they are concatenated and passed into the CNN.}
\label{fig:network-cnn}
\end{figure}

\begin{figure}[t]
\hspace{11mm}
\includegraphics[scale=0.9]{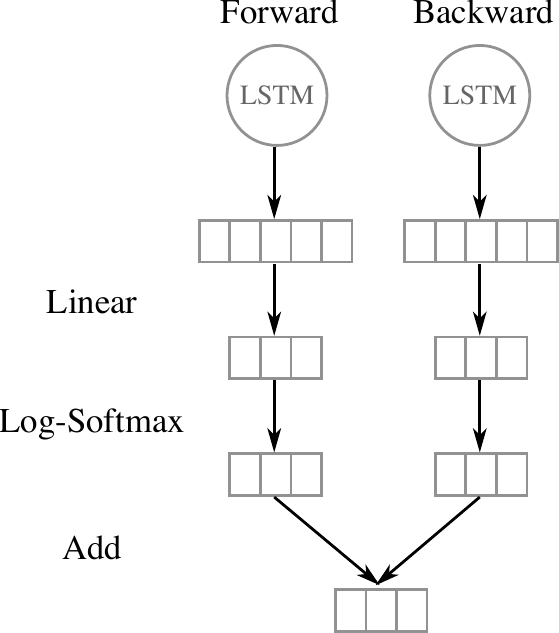} 
\caption{
The output layers (``Out'' in Figure  \ref{fig:network-main}) decode output into a score for each tag category.}
\label{fig:network-out}
\end{figure}

Our neural network is inspired by the work of Collobert et al. \shortcite{collobert2011}, where lookup tables transform discrete features such as words and characters into continuous vector representations, which are then concatenated and fed into a neural network. Instead of a feed-forward network, we use the bi-directional long-short term memory (BLSTM) network. To induce character-level features, we use a convolutional neural network, which has been successfully applied to Spanish and Portuguese NER \cite{dos2015} and German POS-tagging \cite{labeau2015}.

\subsection{Sequence-labelling with BLSTM}

Following the speech-recognition framework outlined by Graves et al. \shortcite{graves2013}, we employed a stacked\footnote{For each direction (forward and backward), the input is fed into multiple layers of LSTM units connected in sequence (i.e. LSTM units in the second layer take in the output of the first layer, and so on); the number of layers is a tuned hyper-parameter. Figure~\ref{fig:network-main} shows only one unit for simplicity.} bi-directional recurrent neural network with long short-term memory units to transform word features into named entity tag scores.  Figures \ref{fig:network-main}, \ref{fig:network-cnn}, and \ref{fig:network-out} illustrate the network in detail.

The extracted features of each word are fed into a forward LSTM network and a backward LSTM network.
The output of each network at each time step is decoded by a linear layer and a log-softmax layer into log-probabilities for each tag category. These two vectors are then simply added together to produce the final output.

We tried minor variants of output layer architecture and selected the one that performed the best in preliminary experiments.

\hspace{-20mm}

\subsection{Extracting Character Features Using a Convolutional Neural Network}

For each word we employ a convolution and a max layer to extract a new feature vector from the per-character feature vectors such as character embeddings (Section \ref{sec:charvec}) and (optionally) character type (Section \ref{sec:char-types}). Words are padded with a number of special {\tt PADDING} characters on both sides depending on the window size of the CNN. 

The hyper-parameters of the CNN are the window size and the output vector size.

\begin{table}[t]
\begin{center}
\begin{tabular}{|l|r|r|}
\hline \bf Category & \bf SENNA & \bf DBpedia \\ \hline
{\tt Location} & 36,697 & 709,772 \\
{\tt Miscellaneous} & 4,722 & 328,575 \\
{\tt Organization} & 6,440 & 231,868 \\
{\tt Person} & 123,283 & 1,074,363 \\ \hline
Total & 171,142 & 2,344,578 \\
\hline
\end{tabular}
\end{center}
\caption{Number of entries for each category in the SENNA lexicon and our DBpedia lexicon.}
\label{tab:lexicon}
\end{table}

\begin{table}[t]
\begin{center}
\small
\begin{tabular}{|l|r|r|r|}
\hline \bf Dataset & \bf Train & \bf Dev & \bf Test \\ \hline
CoNLL-2003 & 204,567 & 51,578 & 46,666\\
& (23,499) & (5,942) & (5,648)\\ \hline
OntoNotes 5.0 & 1,088,503 & 147,724 & 152,728\\
 / CoNLL-2012 & (81,828) & (11,066) & (11,257)\\
\hline
\end{tabular}
\end{center}
\caption{Dataset sizes in number of tokens (entities)}
\label{tab:dataset}
\end{table}

\subsection{Core Features}

\begin{figure*}[ht!]
\begin{center}
\small
\begin{tabular}{c|ccccccccccc}
\small
Text & Hayao & Tada & , & commander & of & the & Japanese & North & China & Area & Army \\ \hline

{\tt LOC}  & {\tt -} & {\tt -} & {\tt -} & {\tt -} & {\tt -} & {\tt B} & {\tt I} & {\tt -} & {\tt S} & {\tt -} & {\tt -} \\
{\tt MISC} & {\tt -} & {\tt -} & {\tt -} & {\tt S} & {\tt B} & {\tt B} & {\tt I} & {\tt S} & {\tt S} & {\tt S} & {\tt S} \\
{\tt ORG}  & {\tt -} & {\tt -} & {\tt -} & {\tt -} & {\tt -} & {\tt B} & {\tt I} & {\tt B} & {\tt I} & {\tt I} & {\tt E} \\
{\tt PERS} & {\tt B} & {\tt E} & {\tt -} & {\tt -} & {\tt -} & {\tt -} & {\tt -} & {\tt -} & {\tt S} & {\tt -} & {\tt -}  
\end{tabular}
\end{center}
\caption{Example of how lexicon features are applied. The {\tt B}, {\tt I}, {\tt E}, markings indicate that the token matches the Begin, Inside, and End token of an entry in the lexicon. {\tt S} indicates that the token matches a single-token entry.}
\label{fig:lexicon-example}
\end{figure*}

\subsubsection{Word Embeddings}
Our best model uses the publicly available 50-dimensional word embeddings released by Collobert et al. \shortcite{collobert2011}\footnote{\url{http://ml.nec-labs.com/senna/}}, which were trained on Wikipedia and the Reuters RCV-1 corpus. 

We also experimented with two other sets of published embeddings, namely Stanford's GloVe embeddings\footnote{\url{http://nlp.stanford.edu/projects/glove/}} trained on 6 billion words from Wikipedia and Web text \cite{pennington2014} and Google's word2vec embeddings\footnote{\url{https://code.google.com/p/word2vec/}} trained on 100 billion words from Google News \cite{mikolov2013}.

In addition, as we hypothesized that word embeddings trained on in-domain text may perform better, we also used the publicly available GloVe \cite{pennington2014} program and an in-house re-implementation\footnote{We used our in-house reimplementation to train word vectors because it uses distributed processing to train much quicker than the publicly-released implementation of word2vec and its performance on the word analogy task was higher than reported by Mikolov et al. \shortcite{mikolov2013}.} of the word2vec \cite{mikolov2013} program to train word embeddings on Wikipedia and Reuters RCV1 datasets as well.\footnote{While Collobert et al. \shortcite{collobert2011} used Wikipedia text from 2007, we used Wikipedia text from 2011.}

Following Collobert et al. \shortcite{collobert2011}, all words are lower-cased before passing through the lookup table to convert to their corresponding embeddings. The pre-trained embeddings are allowed to be modified during training.\footnote{Preliminary experiments showed that modifiable vectors performed better than so-called ``frozen vectors.''}

\subsubsection{Character Embeddings}
\label{sec:charvec}
We randomly initialized a lookup table with values drawn from a uniform distribution with range $[-0.5, 0.5]$ to output a character embedding of 25 dimensions. The character set includes all unique characters in the CoNLL-2003 dataset\footnote{Upper and lower case letters, numbers, and punctuations} plus the special tokens {\tt PADDING} and {\tt UNKNOWN}. The {\tt PADDING} token is used for the CNN, and the {\tt UNKNOWN} token is used for all other characters (which appear in OntoNotes). The same set of random embeddings was used for all experiments.\footnote{We did not experiment with other settings because the English character set is small enough that effective embeddings could be learned directly from the task data.}

\subsection{Additional Word-level Features}

\subsubsection{Capitalization Feature}
\label{sec:caps}
As capitalization information is erased during lookup of the word embedding, we evaluate Collobert's  method of using a separate lookup table to add a capitalization feature with the following options: {\tt allCaps}, {\tt upperInitial}, {\tt lowercase}, {\tt mixedCaps}, {\tt noinfo} \cite{collobert2011}. This method is compared with the character type feature (Section~\ref{sec:char-types}) and character-level CNNs.



\begin{savenotes}
\begin{table*}[t]
\begin{center}
\small
\begin{tabular}{|l|r|c|r|c|}
\hline \multirow{2}{*}{\bf Hyper-parameter} & \multicolumn{2}{c|}{\bf CoNLL-2003 (Round 2)} & \multicolumn{2}{c|}{\bf OntoNotes 5.0 (Round 1)} \\ \cline{2-5}
& Final & Range & Final & Range \\ \hline
Convolution width & \bf 3 & $[3,7]$ & \bf 3 & $[3,9]$ \\
CNN output size & \bf 53 & $[15,84]$ & \bf 20 & $[15,100]$ \\
LSTM state size & \bf 275 & $[100,500]$ & \bf 200 & $[100,400]$\footnote{By increments of 50.}\\
LSTM layers & \bf 1 & $[1,4]$ & \bf 2 & $[2,4]$ \\
Learning rate & \bf 0.0105 & $[10^{-3},10^{-1.8}]$ & \bf 0.008 & $[10^{-3.5},10^{-1.5}]$ \\
Epochs\footnote{Determined by evaluating dev set performance.} & \bf 80 & \-- & \bf 18 & \-- \\
Dropout\footnote{Probability of \emph{discarding} any LSTM output node.} & \bf 0.68 & $[0.25,0.75]$ & \bf \bf 0.63 & $[0,1]$\\
Mini-batch size & \bf 9 & \--\footnote{Mini-batch size was  excluded from the round 2 particle swarm hyper-parameter search space due to time constraints.} & \bf 9 & $[5,14]$ \\
\hline
\end{tabular}
\end{center}
\caption{Hyper-parameter search space and final values used for all experiments}
\label{tab:hyperparams}
\end{table*}
\end{savenotes}

\subsubsection{Lexicons}
Most state of the art NER systems make use of lexicons as a form of external knowledge \cite{ratinov2009,passos2014}. 

For each of the four categories ({\tt Person}, {\tt Organization}, {\tt Location}, {\tt Misc}) defined by the CoNLL 2003 NER shared task, we compiled a list of known named entities from DBpedia \cite{auer2007dbpedia}, by extracting all descendants of DBpedia types corresponding to the CoNLL categories.\footnote{The {\tt Miscellaneous} category was populated by entities of the DBpedia categories {\tt Artifact} and {\tt Work}.} We did not construct separate lexicons for the OntoNotes tagset because correspondences between DBpedia categories and its tags could not be found in many instances. In addition, for each entry we first removed parentheses and all text contained within, then stripped trailing punctuation,\footnote{The punctuation stripped was period, comma, semi-colon, colon, forward slash, backward slash, and question mark.} and finally tokenized it with the Penn Treebank tokenization script for the purpose of partial matching. Table~\ref{tab:lexicon} shows the size of each category in our lexicon compared to Collobert's lexicon, which we extracted from their SENNA system.

Figure \ref{fig:lexicon-example} shows an example of how the lexicon features are applied.\footnote{As can been seen in this example, the lexicons \--- in particular {\tt Miscellaneous} \--- still contain a lot of noise. 
} For each lexicon category, we match every n-gram (up to the length of the longest lexicon entry) against entries in the lexicon. A match is successful when the n-gram matches the prefix or suffix of an entry and is at least half the length of the entry. Because of the high potential for spurious matches, for all categories except {\tt Person}, we discard partial matches less than 2 tokens in length. When there are multiple overlapping matches within the same category, we prefer exact matches over partial matches, and then longer matches over shorter matches, and finally earlier matches in the sentence over later matches. All matches are case insensitive.

For each token in the match, the feature is encoded in BIOES annotation ({\tt Begin}, {\tt Inside}, {\tt Outside}, {\tt End}, {\tt Single}), indicating the position of the token in the matched entry. In other words, {\tt B} will not appear in a suffix-only partial match, and {\tt E} will not appear in a prefix-only partial match.

As we will see in Section~\ref{sec:lexicon-results}, we found that this more sophisticated method outperforms the method presented by Collobert et al. \shortcite{collobert2011}, which treats partial and exact matches equally, allows prefix but not suffix matches, allows very short partial matches, and marks tokens with {\tt YES}/{ \tt NO}.

In addition, since Collobert et al. \shortcite{collobert2011} released their lexicon with their SENNA system, we also applied their lexicon to our model for comparison and investigated using both lexicons simultaneously as distinct features. We found that the two lexicons complement each other and improve performance on the CoNLL-2003 dataset.

Our best model uses the SENNA lexicon with exact matching and our DBpedia lexicon with partial matching, with BIOES annotation in both cases. 

\subsection{Additional Character-level Features}

\label{sec:char-types}
A lookup table was used to output a 4-dimensional vector representing the type of the character (upper case, lower case, punctuation, other).

\begin{table}[t]
\begin{center}
\small
\begin{tabular}{|c|r|r|}
\hline \bf Round & \bf CoNLL-2003 & \bf OntoNotes 5.0 \\ \hline
1 & 93.82 ($\pm$ 0.15) & {\bf 84.57} ($\pm$ 0.27) \\
2 & {\bf 94.03} ($\pm$ 0.23) & 84.47 ($\pm$ 0.29) \\
\hline
\end{tabular}
\end{center}
\caption{Development set F1 score performance of the best hyper-parameter settings in each optimization round.}
\label{tab:hyperopt-result}
\end{table}

\subsection{Training and Inference}

\subsubsection{Implementation}

We implement the neural network using the torch7 library \cite{collobert2011torch7}. Training and inference are done on a per-sentence level. The initial states of the LSTM are zero vectors. Except for the character and word embeddings whose initialization has  been described previously, all lookup tables are randomly initialized with values drawn from the standard normal distribution. 

\subsubsection{Objective Function and Inference}

We train our network to maximize the sentence-level log-likelihood from Collobert et al. \shortcite{collobert2011}.\footnote{Much later, we discovered that training with cross entropy objective while performing Viterbi decoding to restrict output to valid tag sequences also appears to work just as well.}

First, we define a tag-transition matrix $A$ where $A_{i,j}$ represents the score of jumping from tag $i$ to tag $j$ in successive tokens, and $A_{0,i}$ as the score for starting with tag $i$. This matrix of parameters are also learned. Define $\theta$ as the set of parameters for the neural network, and $\theta' = \theta \cup \{A_{i,j} \; \forall i, j\}$ as the set of all parameters to be trained. Given an example sentence, $[x]_1^T$ , of length $T$, and define $[f_{\theta}]_{i,t}$ as the score outputted by the neural network for the $t^{\text{th}}$ word and $i^{\text{th}}$ tag given parameters $\theta$, then the score of a \emph{sequence} of tags $[i]_1^T$ is given as the sum of network and transition scores:
$$S([x]_1^T, [i]_1^T, \theta') = \sum_{t=1}^T \left ( A_{[i]_{t-1}, [i]_t} + [f_{\theta}]_{[i]_t, t} \right )$$


\begin{savenotes}
\begin{table*}[t!]
\begin{center}
\small
\begin{tabular}{|l|c|c|c|c|c|c|}
\hline \bf \multirow{2}{*}{Model} & \multicolumn{3}{c|}{\bf CoNLL-2003} & \multicolumn{3}{c|}{\bf OntoNotes 5.0} \\ \cline{2-7}
 & \bf Prec. & \bf Recall & \bf F1 & \bf Prec. & \bf Recall & \bf F1 \\ \hline
FFNN + emb + caps + lex & 89.54 & 89.80 & 89.67 ($\pm$ 0.24) & 74.28 & 73.61 & 73.94 ($\pm$ 0.43) \\
BLSTM                 & 80.14 & 72.81 & 76.29 ($\pm$ 0.29) & 79.68 & 75.97 & 77.77 ($\pm$ 0.37) \\
BLSTM-CNN             & 83.48 & 83.28 & 83.38 ($\pm$ 0.20) & 82.58 & 82.49 & 82.53 ($\pm$ 0.40) \\
BLSTM-CNN + emb       & 90.75 & 91.08 & 90.91 ($\pm$ 0.20) & 85.99 & 86.36 & 86.17 ($\pm$ 0.22) \\ 
BLSTM-CNN + emb + lex & 91.39 & \bf 91.85 & {\bf 91.62} ($\pm$ 0.33) & \bf 86.04 & \bf 86.53 & {\bf 86.28} ($\pm$ 0.26) \\ \hline

Collobert et al. \shortcite{collobert2011} & - & - & 88.67 & - & - & - \\
Collobert et al. \shortcite{collobert2011} + lexicon & - & - & 89.59 & - & - & - \\
Huang et al. \shortcite{huang2015} & - & - & 90.10 & - & - & - \\ \hline

Ratinov and Roth \shortcite{ratinov2009}\footnote{OntoNotes results taken from \cite{durrett2014}} & 91.20 & 90.50 & 90.80 & 82.00 & 84.95 & 83.45 \\
Lin and Wu \shortcite{lin2009} & - & - & 90.90 & - & - & - \\
Finkel and Manning \shortcite{finkel2009}\footnote{Evaluation on OntoNotes 5.0 done by Pradhan et al. \shortcite{pradhan2013} } & - & - & - & 84.04 & 80.86 & 82.42 \\
Suzuki et al. \shortcite{suzuki2011} & - & - & 91.02 & - & - & - \\
Passos et al. \shortcite{passos2014}\footnote{Not directly comparable as they evaluated on an earlier version of the corpus with a different data split.} & - & - & 90.90 & - & - & 82.24 \\
Durrett and Klein \shortcite{durrett2014} & - & - & - & 85.22 & 82.89 & 84.04 \\
Luo et al. \shortcite{luo2015}\footnote{Numbers taken from the original paper \cite{luo2015}. While the precision, recall, and F1 scores are clearly inconsistent, it is unclear in which way they are incorrect.} & \bf 91.50 & 91.40 & 91.20 & - & - & - \\ \hline
\end{tabular}
\end{center}
\caption{Results of our models, with various feature sets, compared to other published results. The three sections are, in order, our models, published neural network models, and published non-neural network models. For the features, emb~=~Collobert word embeddings, caps~=~capitalization feature, lex~=~lexicon features from both SENNA and DBpedia lexicons. For F1 scores, standard deviations are in parentheses.}
\label{tab:main-result}
\end{table*}
\end{savenotes}

Then, letting $[y]_1^T$ be the true tag sequence, the sentence-level log-likelihood is obtained by normalizing the above score over \emph{all possible tag-sequences} $[j]_1^T$ using a softmax:
\begin{align*}
&\log P([y]_1^T \, | \, [x]_1^T, \theta') \\
&= S([x]_1^T, [y]_1^T, \theta') - \log \sum_{\forall [j]_1^T} e^{S([x]_1^T, [j]_1^T, \theta')}
\end{align*}

This objective function and its gradients can be efficiently computed by dynamic programming \cite{collobert2011}. 

At inference time, given neural network outputs $[f_{\theta}]_{i,t}$ we use the Viterbi algorithm to find the tag sequence $[i]_1^T$ that maximizes the score $S([x]_1^T, [i]_1^T, \theta')$. 

\subsubsection{Tagging Scheme}

The output tags are annotated with BIOES (which stand for {\tt Begin}, 
{\tt Inside}, {\tt Outside}, {\tt End}, {\tt Single}, indicating the position of the token in the entity) as this scheme has been reported to outperform others such as BIO \cite{ratinov2009}.

\subsubsection{Learning Algorithm}

Training is done by mini-batch stochastic gradient descent (SGD) with a fixed learning rate. Each mini-batch consists of multiple sentences with the same number of tokens. 
We found applying dropout to the output nodes\footnote{Adding dropout to inputs seems to have an adverse effect.} of each LSTM layer \cite{pham2014} was quite effective in reducing overfitting (Section~\ref{sec:dropout-results}).
We explored other more sophisticated optimization algorithms such as momentum \cite{nesterov1983}, AdaDelta \cite{zeiler2012}, and RMSProp \cite{hinton2012}, and in preliminary experiments they did not 
    improve upon plain SGD. 

\begin{table*}[ht!]
\begin{center}
\footnotesize
\begin{tabular}{|l|c|c|c|c|c|c|}
\hline 
\multirow{2}{*}{\bf Features} & \multicolumn{2}{|c|}{\bf BLSTM} & \multicolumn{2}{|c|}{\bf BLSTM-CNN} & \multicolumn{2}{|c|}{\bf BLSTM-CNN + lex} \\ \cline{2-7}
& \bf CoNLL & \bf OntoNotes & \bf CoNLL & \bf OntoNotes & \bf CoNLL & \bf OntoNotes\\ \hline
none & 76.29 ($\pm$ 0.29) & 77.77 ($\pm$ 0.37) & 83.38 ($\pm$ 0.20) & 82.53 ($\pm$ 0.40) & 87.77 ($\pm$ 0.29) & 83.82 ($\pm$ 0.19) \\
emb & 88.23 ($\pm$ 0.23) & 82.72 ($\pm$ 0.23) & 90.91 ($\pm$ 0.20) & 86.17 ($\pm$ 0.22) & {\bf 91.62} ($\pm$ 0.33) & 86.28 ($\pm$ 0.26) \\
emb + caps       & 90.67 ($\pm$ 0.16) & 86.19 ($\pm$ 0.25) & 90.98 ($\pm$ 0.18) & 86.35 ($\pm$ 0.28) & 91.55 ($\pm$ 0.19)* & 86.28 ($\pm$ 0.32)* \\
emb + caps + lex & {\bf 91.43} ($\pm$ 0.17) & {\bf 86.21} ($\pm$ 0.16) & {\bf 91.55} ($\pm$ 0.19)* & 86.28 ($\pm$ 0.32)* & 91.55 ($\pm$ 0.19)* & 86.28 ($\pm$ 0.32)* \\ \hline
emb + char & \-- & \-- & 90.88 ($\pm$ 0.48) & 86.08 ($\pm$ 0.40) & 91.44 ($\pm$ 0.23) & {\bf 86.34} ($\pm$ 0.18) \\
emb + char + caps & \-- & \-- & 90.88 ($\pm$ 0.31) & {\bf 86.41} ($\pm$ 0.22) & 91.48 ($\pm$ 0.23) & 86.33 ($\pm$ 0.26) \\ \hline

\end{tabular}
\end{center}
\caption{F1 score results of BLSTM and BLSTM-CNN models with various additional features; emb = Collobert word embeddings, char = character type feature, caps = capitalization feature, lex = lexicon features. Note that starred results are repeated for ease of comparison.}
\label{tab:res-feat-no-cnn}
\end{table*}

\section{Evaluation}

Evaluation was performed on the well-established CoNLL-2003 NER shared task dataset \cite{conll2003} and the much larger but less-studied OntoNotes 5.0 dataset \cite{ontonotes2006,pradhan2013}. Table~\ref{tab:dataset} gives an overview of these two different datasets.

For each experiment, we report the average and standard deviation of 10 successful trials.

\subsection{Dataset Preprocessing}

For all datasets, we performed the following pre-processing:
\begin{itemize}
\item All digit sequences are replaced by a single ``0''.
\item Before training, we group sentences by word length
into mini-batches and shuffle them.
\end{itemize}
In addition, for the OntoNotes dataset, in order to handle the {\tt Date}, {\tt Time}, {\tt Money}, {\tt Percent}, {\tt Quantity}, {\tt Ordinal}, and {\tt Cardinal} named entity tags, we split tokens before and after every digit.

\subsection{CoNLL 2003 Dataset}

The CoNLL-2003 dataset \cite{conll2003} consists of newswire from the Reuters RCV1 corpus tagged with four types of named entities: location, organization, person, and miscellaneous. As the dataset is small compared to OntoNotes, we trained the model on both the training and development sets after performing hyper-parameter optimization on the development set.


\subsection{OntoNotes 5.0 Dataset}

Pradhan et al. \shortcite{pradhan2013} compiled a core portion of the OntoNotes 5.0 dataset for the CoNLL-2012 shared task and described a standard train/dev/test split, which we use for our evaluation. Following Durrett and Klein \shortcite{durrett2014}, we applied our model to the portion of the dataset with gold-standard named entity annotations; the New Testaments portion was excluded for lacking gold-standard annotations. This dataset is much larger than CoNLL-2003 and consists of text from a wide variety of sources, such as broadcast conversation, broadcast news, newswire, magazine, telephone conversation, and Web text.

\subsection{Hyper-parameter Optimization}
\label{sec:hyperopt}

We performed two rounds of hyper-parameter optimization and selected the best settings based on development set performance\footnote{Hyper-parameter optimization was done with the BLSTM-CNN + emb + lex feature set, as it had the best performance.}. Table~\ref{tab:hyperparams} shows the final hyper-parameters, and Table~\ref{tab:hyperopt-result} shows the dev set performance of the best models in each round.  

In the first round, we performed random search and selected the best hyper-parameters over the development set of the CoNLL-2003 data. We evaluated around 500 hyper-parameter settings. Then, we took the same settings and tuned the learning rate and epochs on the OntoNotes development set.
  \footnote{Selected based on dev set performance of a few runs.}

For the second round, we performed independent hyper-parameter searches on each dataset using Optunity's implementation of particle swarm \cite{claesen2014}, as there is some evidence that it is more efficient than random search \cite{clerc2002}. We evaluated 500 hyper-parameter settings this round as well. As we later found out that training fails occasionally (Section~\ref{sec:failed-trials}) as well as large variation from run to run, we ran the top 5 settings from each dataset for 10 trials each and selected the best one based on averaged dev set performance.

For CoNLL-2003, we found that particle swarm produced better hyper-parameters than random search. However, surprisingly for OntoNotes particle swarm was unable to produce better hyper-parameters than those from the ad-hoc approach in round 1. We also tried tuning the CoNLL-2003 hyper-parameters from round 2 for OntoNotes and that was not any better\footnote{The result is 84.41 ($\pm$ 0.33) on the OntoNotes dev set.} either.

We trained CoNLL-2003 models for a large number of epochs because we observed that the models did not exhibit overtraining and instead continued to slowly improve on the development set long after reaching near 100\% accuracy on the training set. In contrast, despite OntoNotes being much larger than CoNLL-2003, training for more than about 18 epochs causes performance on the development set to decline steadily due to overfitting.

\begin{table}[t!]
\begin{center}
\small
\begin{tabular}{|l|c|c|}
\hline \bf Word Embeddings & \bf CoNLL-2003 & \bf OntoNotes \\ \hline
Random 50d       & 87.77 ($\pm$ 0.29) & 83.82 ($\pm$ 0.19) \\
Random 300d      & 87.84 ($\pm$ 0.23) & 83.76 ($\pm$ 0.37) \\
GloVe 6B 50d     & 91.09 ($\pm$ 0.15) & 86.25 ($\pm$ 0.24) \\
GloVe 6B 300d    & 90.71 ($\pm$ 0.21) & 86.26 ($\pm$ 0.30) \\
Google 100B 300d & 90.60 ($\pm$ 0.23) & 85.34 ($\pm$ 0.25) \\ \hline
Collobert 50d    & {\bf 91.62} ($\pm$ 0.33) & {\bf 86.28} ($\pm$ 0.26) \\
Our GloVe 50d    & 91.41 ($\pm$ 0.21) & 86.24 ($\pm$ 0.35) \\
Our Skip-gram 50d & 90.76 ($\pm$ 0.23) & 85.70 ($\pm$ 0.29) \\
\hline
\end{tabular}
\end{center}
\caption{
F1 scores when the Collobert word vectors are replaced. We tried 50- and 300-dimensional random vectors (Random 50d, Random 300d); GloVe's released vectors trained on 6 billion words (GloVe 6B 50d, GloVe 6B 300d); Google's released 300-dimensional vectors trained on 100 billion words from Google News (Google 100B 300d); and 50-dimensional GloVe and word2vec skip-gram vectors that we trained on Wikipedia and Reuters RCV-1 (Our GloVe 50d, Our Skip-gram 50d).
}
\label{tab:wordvec-res}
\end{table}

\subsection{Excluding Failed Trials}
\label{sec:failed-trials}
On the CoNLL-2003 dataset, while BLSTM models completed training without difficulty, the BLSTM-CNN models fail to converge around 5$\sim$10\% of the time depending on feature set. Similarly, on OntoNotes, 1.5\% of trials fail. We found that using a lower learning rate
 reduces failure rate. 
 We also tried clipping gradients and using AdaDelta and both of them were effective at eliminating such failures by themselves.
 AdaDelta, however, made training more expensive with no gain in model performance.

In any case, for all experiments we excluded trials where the final F1 score on a subset of training data falls below a certain threshold, and continued to run trials until we obtained 10 successful ones. 

For CoNLL-2003, we excluded trials where the final F1 score on the development set was less than 95; there was no ambiguity in selecting the threshold as every trial scored either above 98 or below 90. For OntoNotes, the threshold was a F1 score of 80 on the last 5,000 sentences of the training set; every trial scored either above 80 or below 75.

\subsection{Training and Tagging Speed}
On an Intel Xeon E5-2697 processor, training takes about 6 hours while tagging the test set takes about 12 seconds for CoNLL-2003. The times are 10 hours and 60 seconds respectively for OntoNotes.

\begin{table*}[t]
\begin{center}
\small
\begin{tabular}{|c|c|c|c|c|}
\hline \bf \multirow{2}{*}{Dropout} & \multicolumn{2}{c|}{\bf CoNLL-2003} & \multicolumn{2}{c|}{\bf OntoNotes 5.0} \\ \cline{2-5}
 & \bf Dev & \bf Test & \bf Dev & \bf Test \\ \hline
\--     & 93.72 ($\pm$ 0.10) & 90.76 ($\pm$ 0.22) & 82.02 ($\pm$ 0.49) & 84.06 ($\pm$ 0.50) \\
0.10 & 93.85 ($\pm$ 0.18) & 90.87 ($\pm$ 0.31) & 83.01 ($\pm$ 0.39) & 84.94 ($\pm$ 0.25) \\
0.30 & 94.08 ($\pm$ 0.17) & 91.09 ($\pm$ 0.18) & 83.61 ($\pm$ 0.32) & 85.44 ($\pm$ 0.33) \\
0.50 & 94.19 ($\pm$ 0.18) & 91.14 ($\pm$ 0.35) & 84.35 ($\pm$ 0.23) & {\bf 86.36} ($\pm$ 0.28) \\
\textbf{0.63} & \-- & \-- & 84.47 ($\pm$ 0.23) & 86.29 ($\pm$ 0.25) \\
\textbf{0.68} & {\bf 94.31} ($\pm$ 0.15) & {\bf 91.23} ($\pm$ 0.16) & \-- & \-- \\
0.70 & {\bf 94.31} ($\pm$ 0.24) & 91.17 ($\pm$ 0.37) & {\bf 84.56} ($\pm$ 0.40) & 86.17 ($\pm$ 0.25) \\
0.90 & 94.17 ($\pm$ 0.17) & 90.67 ($\pm$ 0.17) & 81.38 ($\pm$ 0.19) & 82.16 ($\pm$ 0.18) \\


\hline
\end{tabular}
\end{center}
\caption{F1 score results with various dropout values. Models were trained using only the training set for each dataset. All other experiments use dropout = 0.68 for CoNLL-2003 and dropout = 0.63 for OntoNotes 5.0.
}
\label{tab:dropout-res-conll}
\end{table*}

\section{Results and Discussion}

Table~\ref{tab:main-result} shows the results for all datasets. To the best of our knowledge, our best models have surpassed the previous highest reported F1 scores for both CoNLL-2003 and OntoNotes. In particular, with no external knowledge other than word embeddings, our model is competitive on the CoNLL-2003 dataset and establishes a new state of the art for OntoNotes, suggesting that given enough data, the neural network automatically learns the relevant features for NER without feature engineering.

\subsection{Comparison with FFNNs}

We re-implemented the FFNN model of Collobert et al. \shortcite{collobert2011} as a baseline for comparison. Table~\ref{tab:main-result} shows that while performing reasonably well on CoNLL-2003, FFNNs are clearly inadequate for OntoNotes, which has a larger domain, showing that LSTM models are essential for NER.

\subsection{Character-level CNNs vs. Character Type and Capitalization Features}

The comparison of models in Table~\ref{tab:res-feat-no-cnn} shows that on CoNLL-2003, BLSTM-CNN models significantly\footnote{Wilcoxon rank sum test, $p < 0.05$ when comparing the four BLSTM models with the corresponding BLSTM-CNN models using the same feature set. The Wilcoxon rank sum test was selected for its robustness against small sample sizes when the distribution is unknown.} outperform the BLSTM models when given the same feature set. This effect is smaller and not statistically significant on OntoNotes when capitalization features are added. 
Adding character type and capitalization features to the BLSTM-CNN models degrades performance for CoNLL and mostly improves performance on OntoNotes, suggesting character-level CNNs can replace hand-crafted character features in some cases, but systems with weak lexicons may benefit from character features.

\subsection{Word Embeddings}

Table~\ref{tab:main-result} and Table~\ref{tab:wordvec-res} show that we obtain a large, significant\footnote{Wilcoxon rank sum test, $p < 0.001$} improvement when trained word embeddings are used, as opposed to random embeddings, regardless of the additional features used. This is consistent with  Collobert et. al. \shortcite{collobert2011}'s results.

Table~\ref{tab:wordvec-res} compares the performance of different word embeddings in our best model in Table~\ref{tab:main-result} (BLSTM-CNN + emb + lex). For CoNLL-2003, the publicly available GloVe and Google embeddings are about one point behind Collobert's embeddings. For OntoNotes, GloVe embeddings perform close to Collobert embeddings while Google embeddings are again one point behind. In addition, 300 dimensional embeddings present no significant improvement over 50 dimensional embeddings \--- a result previously reported by Turian et al. \shortcite{turian2010}.

One possible reason that Collobert embeddings perform better than other publicly available embeddings on CoNLL-2003 is that they are trained on the Reuters RCV-1 corpus, the source of the CoNLL-2003 dataset, whereas the other embeddings are not\footnote{To make a direct comparison to Collobert et al. \shortcite{collobert2011}, we do not exclude the CoNLL-2003 NER task test data from the word vector training data. While it is possible that this difference could be responsible for the disparate performance of word vectors, the CoNLL-2003 training data comprises only 20k out of 800 million words, or 0.00002\% of the total data; in an unsupervised training scheme, the effects are likely negligible.}. On the other hand, we suspect that Google's embeddings perform poorly because of vocabulary mismatch \-- in particular, Google's embeddings were trained in a case-sensitive manner, and embeddings for many common punctuations and symbols were not provided. To test these hypotheses, we performed experiments with new word embeddings trained using GloVe and word2vec, with vocabulary list and corpus similar to Collobert et. al. \shortcite{collobert2011}. As shown in Table~\ref{tab:wordvec-res}, our GloVe embeddings improved significantly\footnote{Wilcoxon rank sum test, $p < 0.01$} over publicly available embeddings on CoNLL-2003, and our word2vec skip-gram  embeddings improved significantly\footnote{Wilcoxon rank sum test, $p < 0.01$} over Google's embeddings on OntoNotes. 

Due to time constraints we did not perform new hyper-parameter searches with any of the word embeddings. As word embedding quality depends on hyper-parameter choice during their training \cite{pennington2014}, and also, in our NER neural network, hyper-parameter choice is likely sensitive to the type of word embeddings used, optimizing them all will likely produce better results and provide a fairer comparison of word embedding quality.

\begin{figure*}[ht]
\begin{center}
\includegraphics[scale=0.7]{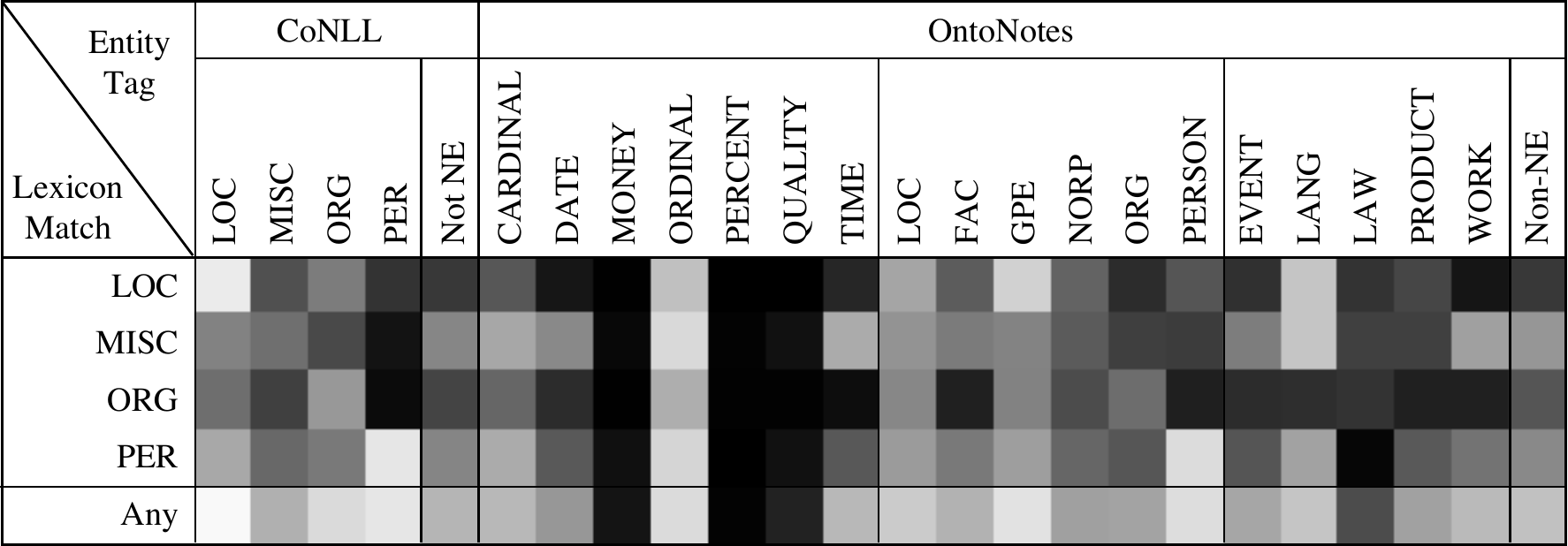} 
\caption{Fraction of named entities of each tag category matched completely by entries in each lexicon category of the SENNA/DBpedia combined lexicon. White = higher fraction.}
\end{center}
\label{fig:lexicon-coverage}
\end{figure*}

\subsection{Effect of Dropout}
\label{sec:dropout-results}

Table~\ref{tab:dropout-res-conll} compares the result of various dropout values for each dataset. The models are trained using only the training set for each dataset to isolate the effect of dropout on both dev and test sets. All other hyper-parameters and features remain the same as our best model in Table~\ref{tab:main-result}. In both datasets and on both dev and test sets, dropout is essential for state of the art performance, and the improvement is statistically significant\footnote{Wilcoxon rank sum test, no dropout vs. best setting: $p < 0.001$ for the CoNLL-2003 test set, $p < 0.0001$ for the OntoNotes 5.0 test set, $p < 0.0005$ for all others. 
}.
Dropout is optimized on the dev set, as described in Section~\ref{sec:hyperopt}. Hence, the chosen value may not be the best-performing in Table~\ref{tab:dropout-res-conll}.

\subsection{Lexicon Features}
\label{sec:lexicon-results}

Table~\ref{tab:res-feat-no-cnn} shows that on the CoNLL-2003 dataset, using features from both the SENNA lexicon and our proposed DBpedia lexicon provides a significant\footnote{Wilcoxon rank sum test, $p < 0.001$.} improvement and allows our model to clearly surpass the previous state of the art. 

\begin{table*}[t]
\begin{center}
\small
\begin{tabular}{|c|c|c|c|c|}
\hline \bf Lexicon & \bf Matching & \bf Encoding & \bf CoNLL-2003 & \bf OntoNotes \\ \hline
No lexicon & - & - & 83.38 ($\pm$ 0.20) & 82.53 ($\pm$ 0.40) \\ \hline
\multirow{2}{*}{SENNA}
 & Exact   & YN    & 86.21 ($\pm$ 0.39) & 83.24 ($\pm$ 0.33) \\ 
 & Exact   & BIOES & 86.14 ($\pm$ 0.48) & 83.01 ($\pm$ 0.52) \\ \hline
\multirow{5}{*}{DBpedia}
 & Exact   & YN    & 84.93 ($\pm$ 0.30) & 83.15 ($\pm$ 0.26) \\
 & Exact   & BIOES & 85.02 ($\pm$ 0.23) & 83.39 ($\pm$ 0.39) \\
 & Partial & YN    & 85.72 ($\pm$ 0.45) & 83.25 ($\pm$ 0.33) \\
 & Partial & BIOES & 86.18 ($\pm$ 0.56) & {\bf 83.97} ($\pm$ 0.38) \\ \cline{2-5}
 & \multicolumn{2}{|c|}{Collobert's method} & 85.01 ($\pm$ 0.31) & 83.24 ($\pm$ 0.26) \\ \hline
Both & \multicolumn{2}{|c|}{Best combination} & {\bf 87.77} ($\pm$ 0.29) & 83.82 ($\pm$ 0.19) \\
\hline
\end{tabular}
\end{center}
\caption{Comparison of lexicon and matching/encoding methods over the BLSTM-CNN model employing random embeddings and no other features. When using both lexicons, the best combination of matching and encoding is Exact-BIOES for SENNA and Partial-BIOES for DBpedia. Note that the SENNA lexicon already contains ``partial entries'' so exact matching in that case is really just a more primitive form of partial matching.}
\label{tab:lexicon-res}
\end{table*}

Unfortunately the difference is minuscule for OntoNotes, most likely because our lexicon does not match DBpedia categories well. Figure 5 shows that on CoNLL-2003, lexicon coverage is reasonable and matches the tags set for everything except the catch-all \texttt{MISC} category. For example, \texttt{LOC} entries in lexicon match mostly \texttt{LOC} named entities and vice versa. However, on OntoNotes, the matches are noisy and correspondence between lexicon match and tag category is quite ambiguous. For example, all lexicon categories have spurious matches in unrelated named entities like \texttt{CARDINAL}, and \texttt{LOC}, \texttt{GPE}, and \texttt{LANGUAGE} entities all get a lot of matches from the \texttt{LOC} category in the lexicon. In addition, named entities in categories like \texttt{NORP}, \texttt{ORG}, \texttt{LAW}, \texttt{PRODUCT} receive little coverage. The lower coverage, noise, and ambiguity all contribute to the disappointing performance. This suggests that the DBpedia lexicon construction method needs to be improved. A reasonable place to start would be the DBpedia category to OntoNotes NE tag mappings.

In order to isolate the contribution of each lexicon and matching method, we compare different sources and matching methods on a BLSTM-CNN model with randomly initialized word embeddings and no other features or sources of external knowledge. Table~\ref{tab:lexicon-res} shows the results. In this weakened model, both lexicons contribute significant\footnote{Wilcoxon rank sum test, $p < 0.05$ for SENNA-Exact-BIOES, $p < 0.005$ for all others.} improvements over the baseline.

Compared to the SENNA lexicon, our DBpedia lexicon is noisier but has broader coverage, which explains why when applying it using the same method as Collobert et al. \shortcite{collobert2011}, it performs worse on CoNLL-2003 but better on OntoNotes -- a dataset containing many more obscure named entities.
However, we suspect that the method of Collobert et al. \shortcite{collobert2011} is not noise resistant and therefore unsuitable for our lexicon because it fails to distinguish exact and partial matches\footnote{We achieve this by using BIOES encoding and prioritizing exact matches over partial matches.} and does not set a minimum length for partial matching.\footnote{Matching only the first word of a long entry is not very useful; this is not a problem in the SENNA lexicon because 99\% of its entries contain only 3 tokens or less.}
Instead, when we apply our superior partial matching algorithm and BIOES encoding with our DBpedia lexicon, we gain a significant\footnote{Wilcoxon rank sum test, $p < 0.001$.} improvement, allowing our lexicon to perform similarly to the SENNA lexicon. 
Unfortunately, as we could not reliably remove partial entries from the SENNA lexicon, we were unable to investigate whether or not our lexicon matching method would help in that lexicon. 

In addition, using both lexicons together as distinct features provides a further improvement\footnote{Wilcoxon rank sum test, $p < 0.001$.} on CoNLL-2003, which we suspect is because the lexicons are complementary; the SENNA lexicon is relatively clean and tailored to newswire, whereas the DBpedia lexicon is noisier but has high coverage.

\subsection{Analysis of OntoNotes Performance}

Table~\ref{tab:ner-genre} shows the per-genre breakdown of the OntoNotes results. As expected, our model performs best on clean text like broadcast news (BN) and newswire (NW), and worst on noisy text like telephone conversation (TC) and Web text (WB). Our model also substantially improves over previous work on all genres except TC, where the small size of the training data likely hinders learning. Finally, the performance characteristics of our model appear to be quite different than the previous CRF models \cite{finkel2009,durrett2014}, likely because we apply a completely different machine learning method.

\begin{savenotes}
\begin{table*}[t]
\small
\begin{center}
\begin{tabular}{|l|r|r|r|r|r|r|}
\hline \bf Model & \bf BC & \bf BN & \bf MZ & \bf NW & \bf TC & \bf WB \\ \hline
Test set size (\# tokens)    & 32,576 & 23,557 & 18,260 & 51,667 & 11,015 & 19,348 \\
Test set size (\# entities)  &  1,697 &  2,184 &  1,163 &  4,696 &    380 &  1,137 \\ \hline
Finkel and Manning \shortcite{finkel2009} & 78.66 & 87.29 & 82.45 & 85.50 & 67.27 & 72.56 \\ 
Durrett and Klein \shortcite{durrett2014}\footnote{We downloaded their publicly released software and model to perform the per-genre evaluation.} & 78.88 & 87.39 & 82.46 & 87.60 & \bf 72.68 & 76.17 \\ \hline
BLSTM-CNN & 81.26 & 86.87 & 79.94 & 85.27 & 67.82 & 72.11 \\ 
BLSTM-CNN + emb & 85.05 & \bf 89.93 & 84.31 & 88.35 & 72.44 & 77.90\\ 
BLSTM-CNN + emb + lex & \bf 85.23 & \bf 89.93 & \bf 84.45 & \bf 88.39 & 72.39 & \bf 78.38\\
\hline
\end{tabular}
\end{center}
\caption{Per genre F1 scores on OntoNotes. BC = broadcast conversation, BN = broadcast news, MZ = magazine, NW = newswire, TC = telephone conversation, WB = blogs and newsgroups}
\label{tab:ner-genre}
\end{table*}
\end{savenotes}

\section{Related Research}

Named entity recognition is a task with a long history. In this section, we summarize the works we compare with and that influenced our approach.

\subsection{Named Entity Recognition}

Most recent approaches to NER have been characterized by the use of CRF, SVM, and perceptron models, where performance is heavily dependent on feature engineering. Ratinov and Roth \shortcite{ratinov2009} used non-local features, a gazetteer extracted from Wikipedia, and Brown-cluster-like word representations, and achieved an F1 score of 90.80 on CoNLL-2003. Lin and Wu \shortcite{lin2009} surpassed them without using a gazetteer by instead using phrase features obtained by performing k-means clustering over a private database of search engine query logs. Passos et al. \shortcite{passos2014} obtained nearly the same performance using only public data by training phrase vectors in their lexicon-infused skip-gram model. In order to combat the problem of sparse features, Suzuki et al. \shortcite{suzuki2011} employed large-scale unlabelled data to perform feature reduction and achieved an F1 score of 91.02 on CoNLL-2003, which is the current state of the art for systems without external knowledge.

Training an NER system together with related tasks such as entity linking has recently been shown to improve the state of the art. Durrett and Klein \shortcite{durrett2014} combined coreference resolution, entity linking, and NER into a single CRF model and added cross-task interaction factors. Their system achieved state of the art results on the OntoNotes dataset, but they did not evaluate on the CoNLL-2003 dataset due to lack of coreference annotations. Luo et al. \shortcite{luo2015} achieved state of the art results on CoNLL-2003 by training a joint model over the NER and entity linking tasks, the pair of tasks whose inter-dependencies contributed the most to the work of Durrett and Klein \shortcite{durrett2014}.

%

\subsection{NER with Neural Networks}

While many approaches involve CRF models, there has also been a long history of research involving neural networks. Early attempts were hindered by lack of computational power, scalable learning algorithms, and high quality word embeddings. 

Petasis et al. \shortcite{petasis2000} used a feed-forward neural network with one hidden layer on NER and achieved state-of-the-art results on the MUC6 dataset. Their approach used only POS tag and gazetteer tags for each word, with no word embeddings. 

Hammerton \shortcite{hammerton2003} attempted NER with a single-direction LSTM network and a combination of 
word vectors trained using self-organizing maps and context vectors obtained using principle component analysis. However, while our method optimizes log-likelihood and uses softmax, they used a different output encoding and optimized an unspecified objective function. Hammerton's \shortcite{hammerton2003} reported results were only slightly above baseline models. 

Much later, with the advent of neural word embeddings, Collobert et al. \shortcite{collobert2011} presented SENNA, which employs a deep FFNN and word embeddings to achieve near state of the art results on POS tagging, chunking, NER, and SRL. We build on their approach, sharing the word embeddings, feature encoding method, and objective functions.

Recently, Santos et al. \shortcite{dos2015} presented their CharWNN network, which augments the neural network of Collobert et al. \shortcite{collobert2011} with character level CNNs, and they reported improved performance on Spanish and Portuguese NER. We have successfully incorporated character-level CNNs into our model. 

There have been various other similar architecture proposed for various sequential labeling NLP tasks.
Huang et al. \shortcite{huang2015} used a BLSTM for the POS-tagging, chunking, and NER tasks, but they employed heavy feature engineering instead of using a CNN to automatically extract character-level features. 
Labeau et al. \shortcite{labeau2015} used a BRNN with character-level CNNs to perform German POS-tagging; our model differs in that we use the more powerful LSTM unit, which we found to perform better than RNNs in preliminary experiments, and that we employ word embeddings, which is much more important in NER than in POS tagging.
Ling et al. \shortcite{ling2015} used both word- and character-level BLSTMs to establish the current state of the art for English POS tagging. While using BLSTMs instead of CNNs allows extraction of more sophisticated character-level features, we found in preliminary experiments that for NER it did not perform significantly better than CNNs and was substantially more computationally expensive to train.

\section{Conclusion}

We have shown that our neural network model, which incorporates a bidirectional LSTM and a character-level CNN and which benefits from robust training through dropout, achieves state-of-the-art results in named entity recognition with little feature engineering. Our model improves over previous best reported results on two major datasets for NER, 
suggesting that the model is capable of learning complex relationships from large amounts of data.

Preliminary evaluation of our partial matching lexicon algorithm suggests that performance could be further improved through more flexible application of existing lexicons. Evaluation of existing word embeddings suggests that the domain of training data is as important as the training algorithm. 

More effective construction and application of lexicons and word embeddings are areas that require more research. In the future, we would also like to extend our model to perform similar tasks such as extended tagset NER and entity linking.


\section*{Acknowledgments}

This research was supported by Honda Research Institute Japan Co., Ltd. The authors would like to thank Collobert et al. \shortcite{collobert2011} for releasing SENNA with its word vectors and lexicon, the torch7 framework contributors, and Andrey Karpathy for the reference LSTM implementation.


\bibliographystyle{acl}
\bibliography{references}

\end{document}